\documentclass[conference]{IEEEtran}
\usepackage{cite}
\usepackage{amsmath,amssymb,amsfonts}
\usepackage{algorithmic}
\usepackage{graphicx}
\usepackage{textcomp}
\usepackage{xcolor}
\usepackage{pifont}
\IEEEoverridecommandlockouts
\begin{document}

\title{Human Kinematics-inspired Skeleton-based Video Anomaly Detection}

\author{\IEEEauthorblockN{Jian Xiao}
\IEEEauthorblockA{\textit{School of Computer and Electronic} \\
\textit{Information /Artificial Intelligence} \\
\textit{Nanjing Normal University}\\
Nanjing, China \\
212202033@njnu.edu.cn}
\and
\IEEEauthorblockN{Tianyuan Liu}
\IEEEauthorblockA{\textit{Department of Industrial} \\
\textit{and Systems Engineering} \\
\textit{The Hong Kong Polytechnic University}\\
Hong Kong, China \\
tianyuan.liu@polyu.edu.hk}
\and
\IEEEauthorblockN{Genlin Ji*}
\IEEEauthorblockA{\textit{School of Computer and Electronic} \\
\textit{Information /Artificial Intelligence} \\
\textit{Nanjing Normal University}\\
Nanjing, China \\
glji@njnu.edu.cn}
}

\maketitle

\begin{abstract}
Previous approaches to detecting human anomalies in videos have typically relied on implicit modeling by directly applying the model to video or skeleton data, potentially resulting in inaccurate modeling of motion information. In this paper, we conduct an exploratory study and introduce a new idea called HKVAD (Human Kinematic-inspired Video Anomaly Detection) for video anomaly detection, which involves the explicit use of human kinematic features to detect anomalies. To validate the effectiveness and potential of this perspective, we propose a pilot method that leverages the kinematic features of the skeleton pose, with a specific focus on the walking stride, skeleton displacement at feet level, and neck level. Following this, the method employs a normalizing flow model to estimate density and detect anomalies based on the estimated density. Based on the number of kinematic features used, we have devised three straightforward variant methods and conducted experiments on two highly challenging public datasets, ShanghaiTech and UBnormal. Our method achieves good results with minimal computational resources, validating its effectiveness and potential.
\end{abstract}

\begin{IEEEkeywords}
video anomaly detection, divide and conquer, deep learning, multi-modality
\end{IEEEkeywords}

\section{Introduction} \label{sec:introduction}
Video anomaly detection aims to identify anomalous events in videos that significantly differ from the normal data distribution \cite{memae}. The task is challenging due to the infrequent occurrence of anomalous events and the context-dependent nature of what constitutes an anomaly \cite{MNAD}. For example, driving on a pedestrian walkway is an anomalous event, while driving a car on a highway is normal. Researchers face challenges in gathering a sufficient number of anomaly samples for supervised model training \cite{future}. Therefore, anomaly detection problems are often approached as one-class classification problems \cite{gold}, using only normal data for model training, and considering data far from the normal pattern as anomalous during testing \cite{cai2021appearance}.

Methods based on video data \cite{multi-task,hybrid,jigsaw,liu2023diversity} have been extensively investigated; however, due to their requirement for substantial computing resources and a lack of interpretability, they are unable to fulfill practical application requirements. In contrast, skeleton-based methods \cite{normalgraph,graphembed,learning,multitimescale,Explicit} have attracted attention from researchers due to the compact and structured nature of skeleton data. This approach extracts human skeleton key points from videos to convert video data into a skeleton-based form, and then uses machine learning models to directly perform anomaly detection on skeleton data. Examples include reconstructing or predicting skeleton sequences \cite{normalgraph,learning,multitimescale,Explicit}, and clustering encoded pose features \cite{graphembed}.

\begin{figure*}[htbp]
\includegraphics[width=\linewidth]{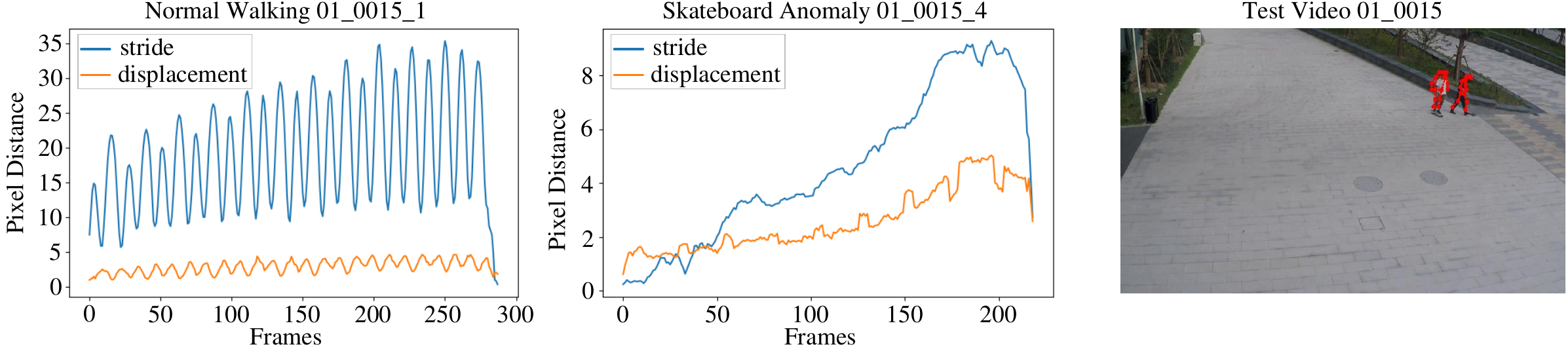}
\caption{The left image shows that during normal walking, the stride displays periodic changes with displacement. However, during skateboarding (middle image), the stride lacks periodic changes, indicating that it is not normal walking. On the right, in scenarios from the ShanghaiTech dataset \cite{shanghaitech}, when a person maintains a standing posture while riding a skateboard, method \cite{learning} misidentifies it as normal walking.}
\label{motivation}
   \vspace{-1em}
\end{figure*}

However, these approaches implicitly model motion information by directly applying neural networks to skeleton data, without incorporating the guiding principles of physical laws and kinematic features. This deficiency may potentially result in inaccurate modeling of human motion, consequently impacting the model's capability for anomaly detection, as supported by prior research \cite{learning}. Specifically, as shown in the right image of Figure \ref{motivation}, the model erroneously classifies an individual maintaining an upright posture while skateboarding (considered an anomalous event in the ShanghaiTech dataset \cite{shanghaitech}) as normal walking, highlighting its inability to accurately capture the complexities of human motion modeling.

Human motion, while highly complex, adheres to physical laws and can be described using kinematic and dynamic equations rooted in classical mechanics theory. This realization prompts us: perhaps employing kinematic features could offer a more precise characterization of human motion and a more efficient detection of abnormal movements. Therefore, We have undertaken preliminary endeavors in this direction, as depicted in the left and middle images of Figure \ref{motivation}. While examining the kinematic features of stride and displacement at feet level, it becomes evident that regular walking is marked by cyclic variations in stride length (i.e., the distance between both feet while walking), whereas skateboarding lacks this characteristic. However, the existing models, such as \cite{learning} incorrectly classify this skateboarding as normal behavior. Therefore, if the model can recognize the differences between these two behaviors in terms of kinematic features, it may be more likely to accurately detect abnormalities.

In this paper, we propose a methodology known as HKVAD, which aims to leverage the kinematic features of skeleton pose data for the purpose of modeling human motion, with subsequent application in anomaly detection. Due to the large number of kinematic features related to human actions, it is not possible to consider all of them. Hence, to validate the effectiveness and potential of the HKVAD, we construct three simple variants of HKVAD based on the number of features used. First, we extract human kinematic features from skeletal data and organize them into time series data. Then, we use a normalizing flow model to estimate the density and detect anomalies in this obtained time series data. Experimental results on two challenging datasets have demonstrated the effectiveness and potential of HKVAD. Additionally, we design a new and more challenging anomaly detection setting that includes rare normal events, which is discussed in Section \ref{con:ShanghaiTech}.



\section{Related Work}

\subsection{Video Anomaly Detection}
From the perspective of the data used, existing video anomaly detection methods can be divided into two categories: methods based on video data and methods based on skeleton data. In current research, methods based on video data typically dominate and commonly employ neural networks to directly capture appearance and motion information from sequences of video frames (or spatiotemporal cubes \cite{multi-task,jigsaw} generated by cropping video frames). In contrast, methods based on skeleton modalities have not been widely explored, but have received increasing attention in recent years.

MPED-RNN \cite{learning} was the first method to propose the use of skeleton data to address the problem of anomaly detection in videos. It learns normal motion patterns by decoupling the reconstruction and prediction of key joints, using RNNs. In contrast, \cite{normalgraph} proposed using graph convolutional networks (GCNs) to model skeleton data. Subsequently, GEPC \cite{graphembed} modeled pose graphs using an autoencoder to learn motion representations and then used clustering methods to obtain normal scores. To capture anomalies at different time scales, \cite{multitimescale} designed a multi-scale prediction framework to model human pose trajectories. To extract motion features from pose sequences, \cite{Explicit} proposed the Motion Embedder from a probabilistic perspective to provide representations of pose motion and applied Transformer to reconstruct pose sequences.

We acknowledge that there is a precedent for introducing physical concepts such as velocity (the displacement between adjacent frames) into the field of video anomaly detection, such as in reference \cite{Explicit}. This approach calculates the skeleton's center point, calculates velocity based on it, and models these velocities probabilistically. However, it only models velocities statistically, without fully exploring the kinematic information of human motion.

\subsection{Normalizing Flows}
\label{con:NormalizingFlows}
Let $x \in\mathbb{R}^d$ be a d-dimensional random variable. A normalizing flow is to transform a complex probability distribution into a simpler one, typically a standard normal distribution, by applying a series of invertible mappings. Mathematically, a normalizing flow consists of a sequence of invertible transformations $f_i: \mathbb{R}^d \to \mathbb{R}^d$, where $i = 1, \dots, K$, such that the composition $f = f_K \circ f_{K-1} \circ \dots \circ f_1$ is also invertible. Let $z=f(x)$ with probability
density function $q(z)$. Through the change of variables formula, we can express the density of $x$:
\begin{equation}p(x)=q(f(x))\left|\det\left(\frac{\partial f}{\partial x}\right)\right|\tag{1}\end{equation}
The determinant of the Jacobian matrix measures how much the volume of the transformed space changes under the transformation $f$. Using logarithms and the chain rule yields the following equation:
\begin{equation}\log p(x)=\log q(f(x))+\sum_{i=1}^{K}\log\left|\det\left(\frac{\partial f_i}{\partial f_{i-1}}\right)\right|\tag{2}\label{con:2}\end{equation}
To compute the density function $p(x)$ in practical applications, the Jacobian determinant in equation (\ref{con:2}) must be computationally feasible.

Normalizing flows have gained widespread popularity in density estimation, with representative methods including NICE \cite{nice}, RealNVP \cite{rnvp}, MAF \cite{maf}, and Glow \cite{glow}. However, normalizing flows have been explored very little in video anomaly detection. \cite{implict} proposed ITAE by using normalized flows to estimate the likelihood of static and dynamic features obtained from RGB data.

\begin{figure*}[htbp]
\includegraphics[width=\linewidth]{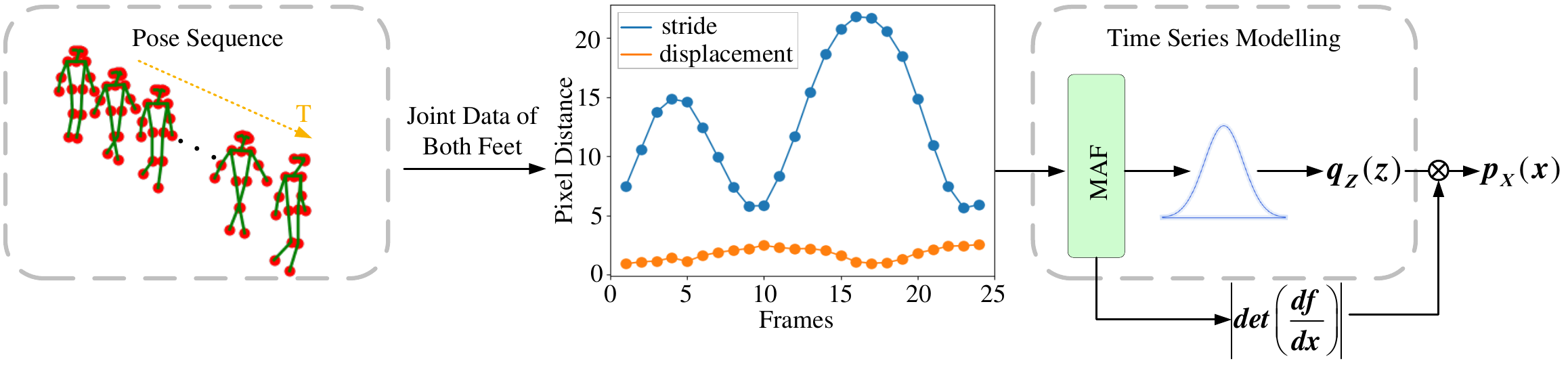}
\caption{An overview of HKVAD-2: First, the joint data from both feet are used to obtain stride-time series and displacement-time series. Then, both time series are simultaneously input into a normalizing flow model, MAF \cite{maf}, for density estimation.}
\label{framework}
\end{figure*}

\section{Method}
To validate the effectiveness and potential of the HKVAD, we opted for the utilization of simplest kinematic features such as displacement and stride, while constraining the amount of data used (no more than three joint data) and employing uncomplicated model architectures. This chosen strategy is based on the assumption that if the model performs well under these stringent constraints, it implies that it is likely to exhibit superior performance when equipped with more extensive data resources and sophisticated model designs, thereby confirming HKVAD's effectiveness and potential. Specifically, we designed three variants: HKVAD-1, which utilizes only displacement-time sequences; HKVAD-2, which utilizes both displacement-time and stride-time sequences; and HKVAD-3, which utilizes displacement-time, stride-time, and neck displacement-time sequences. An overview of HKVAD-2 is shown in Figure \ref{framework}.

\subsection{Data Preprocessing}
We assume that the skeleton has been extracted from the video and the skeleton sequence is formed by the tracking algorithm, then the skeleton sequence of a person in the video can be represented as $X_i\in \mathbb{R}^{T\times k\times 2}$, where $T$ represents the time length of the sequence, $k$ represents the number of joints in the skeleton, and 2 represents the two coordinates, $x$ and $y$, used to represent each joint. Subsequently, we will employ a single subject as a case study to illustrate the methodology of transforming skeleton data into time-series data with physical significance.

\textbf{Stride-Time Series}
At time t, the stride, which is the distance between the two feet during walking, can be computed using the formula that involves the coordinates of the left foot $(x_{i,t}^{left},y_{i,t}^{left})$ and the right foot $(x_{i,t}^{right},y_{i,t}^{right})$:
\begin{equation}s_i^t = \sqrt{(x_{i,t}^{right} - x_{i,t}^{left})^2 + (y_{i,t}^{right} - y_{i,t}^{left})^2}\tag{3}\end{equation}
Subsequently, the stride-time sequence $S_i=\{s_i^1,s_i^2,...,s_i^T\}\in \mathbb{R}^{T\times 1}$ can be obtained.

\textbf{Displacement-Time Series}
In this paper, displacement is defined as the movement between adjacent timestamps (i.e. adjacent frames), which is also computed from the data collected from both feet. Firstly, the coordinates of the midpoint between the two feet are calculated at each time point using the following formula:
\begin{equation}x_{i,t}^{middle} = \frac{x_{i,t}^{right}+x_{i,t}^{left}}{2},\tag{4}\end{equation}
\begin{equation}y_{i,t}^{middle} = \frac{y_{i,t}^{right}+y_{i,t}^{left}}{2}\tag{5}\end{equation}
Then, the displacement is calculated using the following formula:

\begin{small}\begin{equation}d_i^t=\sqrt{(x_{i,t}^{middle}-x_{i,t-1}^{middle})^2+(y_{i,t}^{middle}-y_{i,t-1}^{middle})^2}\tag{6}\end{equation}\end{small}
Wherein it is specifically supplemented that $d_i^1=0$, and then the displacement-time sequence $D_i=\{d_i^1,d_i^2,...,d_i^T\}\in \mathbb{R}^{T\times 1}$ is obtained.

\textbf{Neck Displacement-Time Series}
We use neck displacement to roughly simulate the change of the human body's center of mass, which can be calculated from neck joint data using the following formula:
\begin{equation}n\_d_i^t = \sqrt{(x_{i,t}^{neck} - x_{i,t-1}^{neck})^2 + (y_{i,t}^{neck} - y_{i,t-1}^{neck})^2}\tag{7}\end{equation}
This generates the neck displacement-time series $N\_D_i=\{n\_d_i^1,n\_d_i^2,\dots,n\_d_i^T\}\in \mathbb{R}^{T\times 1}$, where $n\_d_i^1$ is set to 0.

\textbf{Remove Outliers and Smooth Curve}
Due to the influence of random fluctuations in the posture estimation process, the time series calculated using the method above may inevitably contain outliers. Therefore, we first use the $3\sigma$ rule to remove outliers, replacing them with 0. Then, we use a moving sliding average to smooth the time series data, according to the following formula:
\begin{equation}y_t = \frac{1}{2w+1}\sum_{j=-w}^wx_{t+j}\tag{8}\end{equation}
where $x_t$ represents the original data point, $y_t$ represents the smoothed value, $w$ is the window size. Then we obtain the sequences $S_{i}^{'}, D_{i}^{'}, N\_D_{i}^{'} \in \mathbb{R}^{T \times 1}$.

\subsection{Time Series Modelling}
Taking HKVAD-2 as an example, the process of temporal modeling can be described as follows. After obtaining the temporal data $S_{i}^{'}$ and $D_{i}^{'}$, they are concatenated to form $SD_i=[S_{i}^{'},D_{i}^{'}] \in \mathbb{R}^{T \times 2}$. Then, $SD_i$ is segmented into fixed-length temporal segments  using a sliding window method. These temporal segments are then subjected to density estimation using a normalized flow model. The specific choice of normalized flow model can be any of the models proposed in the literature, such as RealNVP or MAF, but in this case, we use MAF. The details of the method are not discussed here as they are not the focus of this article.

We train the model by minimizing the negative log-likelihood function:
\begin{equation}L= - \log q(f(x)) - \log \left|\det\left(\frac{\partial f}{\partial x}\right)\right|\tag{9}\label{con:8}\end{equation}

Let $z \sim \mathcal{N}(\mu, {I})$, then formula (\ref{con:8}) can be further written as follows:
\begin{equation}L= \frac{1}{2}(f(x) - \mu)^2  -Const - \log \left|\det\left(\frac{\partial f}{\partial x}\right)\right|\tag{10}\end{equation}

\subsection{Normality Score}
During the inference process, we score each temporal segment separately. For video clips involving more than one person, we take the minimum score across all individuals in a given frame:
\begin{equation}score = \min_{\forall x\in X^t} \log p(x)\tag{11}\end{equation}
$X^t$ represents the set of temporal segments involved at time $t$.

\section{Experiments}

\subsection{Datasets}
\label{con:ShanghaiTech}
We evaluate out method on two datasets for video anomaly detection, ShanghaiTech \cite{shanghaitech} and UBnormal \cite{ubnormal}, and use Micro-AUC \cite{AUC} as the evaluation metric.

\begin{figure}[t]
  \centering
   \includegraphics[width=\linewidth]{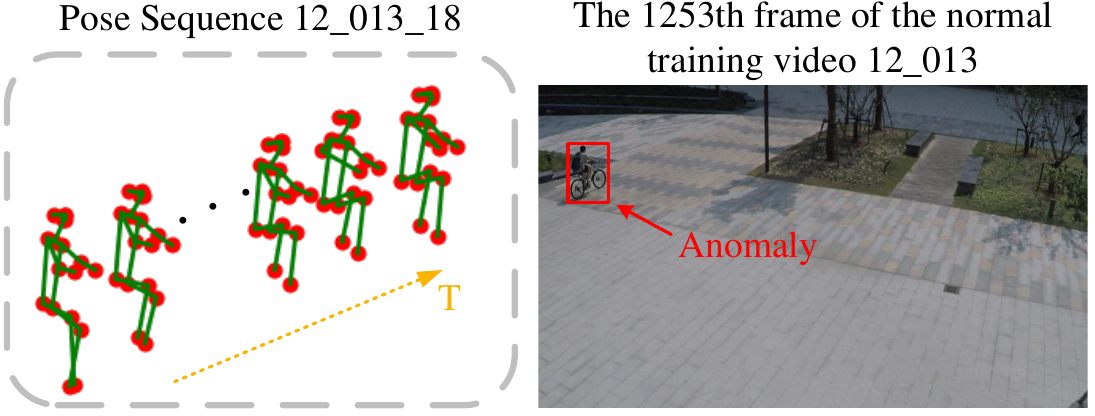}
   \vspace{-1em}
   \caption{
   The left image shows the visualization of the skeleton sequence 12\_013\_18 on the ShanghaiTech training set, while the right image shows a particular frame of the corresponding video scene.}
   \label{fig:dataset}
\end{figure}

\noindent
{\bf ShanghaiTech} \cite{shanghaitech}
is one of the widely used anomaly detection datasets, which includes 330 training videos and 107 testing videos covering 13 different scenarios. The testing videos include anomalous behaviors related to humans (such as jumping and riding bicycles) as well as non-human related anomalies (such as the appearance of large vehicles). ShanghaiTech-HR is a subset of ShanghaiTech dataset that excludes anomalies unrelated to humans. Although the official released training set should only contain normal videos, we have noticed that 12\_013 and 12\_014 mistakenly include anomalies, and the current methods do not exclude the anomalies in the training set during the training phase, as shown in Figure \ref{fig:dataset}. We propose two solutions: 1) we recommend deleting these two erroneous videos to correct the dataset. however, in this paper, to ensure a fair comparison, our experiments were conducted on datasets that still include erroneous data. 2) \textbf{New Anomaly Detection Setting}: we suggest re-annotating the test set and introducing a more challenging setting for video anomaly detection, including rare normal events. That is to say, abnormal behaviors such as riding bicycles, tricycles, or jogging that appear in the training set can actually be considered as rare normal events. Limited by space, we cannot discuss this further.


\begin{table*}[h]
  \begin{center}
    \caption{Comparing with state-of-the-art methods based on AUC score, model parameters, training time, and testing time per epoch on a single CPU, with bold highlighting methods related to this work. In the ShanghaiTech comparison, all compared methods used skeleton data.}
    \resizebox{\textwidth}{!}{
    \begin{tabular}{c c c c c c}
    \hline
      \text{Method} & \text{ShanghaiTech} & \text{ShanghaiTech-HR} & \text{Parameters} & \text{Training Time} & \text{Testing Time}\\
      \hline
      MPED-RNN \cite{learning} & 73.4 & 75.4 & 25.464K & 519s & 34s\\
      MTTP \cite{multitimescale} & 76.03 & 77.04 & - & - & -\\
      GEPC \cite{graphembed} & 76.1 & - & 1598.382K & \textgreater1h & -\\
      Normal Graph \cite{normalgraph} & 74.1 & 76.5 & - & - & -\\
      MoPRL \cite{Explicit} & 81.26 & 82.38 & - & - & -\\
      MSTA-GCN \cite{multiscale} & 75.9 & 76.8 & - & - & -\\
      STG-NF \cite{hirschorn2022normalizing} & 85.937 & 87.375 & 0.616K & 115s & -\\
      \textbf{HKVAD-1(ours)} & \textbf{71.688} & \textbf{72.296} & \textbf{0.027K} & \textbf{\textless6s} & \textbf{\textless3s}\\
      \textbf{HKVAD-2(ours)} & \textbf{67.267} & \textbf{67.675} & \textbf{0.042K} & \textbf{\textless6s} & \textbf{\textless3s}\\
      \textbf{HKVAD-3(ours)} & \textbf{72.206} & \textbf{73.150} & \textbf{0.057K} & \textbf{\textless7s} & \textbf{\textless3s}\\
      \hline
    \end{tabular}}
    \label{ShanghaiTechTable}  
  \end{center}
\end{table*}

\begin{table*}[t]
  \begin{center}
    \caption{Comparison with state-of-the-art methods based on data modality, AUC score, model parameters, training time, and testing time per epoch on a single CPU, with bold highlighting methods related to this work.}
    \resizebox{\textwidth}{!}{
    \begin{tabular}{c c c c c c}
    \hline
    \text{Method} & \text{Data Modality} & \text{UBnormal} & \text{Parameters} & \text{Training Time} & \text{Testing Time}\\
      \hline
      MIL \cite{real} & video & 50.3 & - & - & -\\
      GEPC \cite{graphembed} & skeleton & 52.0 & 1598.382K & - & -\\
      Jigsaw\textbf{*} \cite{jigsaw} & video & 55.57 & 1586.946k & \textgreater1h & \textgreater1h\\
      SSMTL++ \cite{ssmtl++} & video & 62.1 & - & - & -\\
      STG-NF \cite{hirschorn2022normalizing} & skeleton & 71.783 & 0.616K & 122s & -\\
      \textbf{HKVAD-1 (ours)} & \textbf{skeleton} & \textbf{57.240} & \textbf{0.018K} & \textbf{\textless2s} & \textbf{\textless5s}\\
      \textbf{HKVAD-2 (ours)} & \textbf{skeleton} & \textbf{58.667} & \textbf{0.028K} & \textbf{\textless2s} & \textbf{\textless5s}\\
      \textbf{HKVAD-3 (ours)} & \textbf{skeleton} & \textbf{59.573} & \textbf{0.038K} & \textbf{\textless2s} & \textbf{\textless5s}\\
      \hline
      \multicolumn{4}{l}{$^{*}$denotes that the results in the table are from our implementation.}
    \end{tabular}
    }
    \label{UBnormalTable}  
  \end{center}
\end{table*}

\noindent
{\bf UBnormal} \cite{ubnormal}
is one of the most challenging datasets proposed in recent years, containing 22 types of anomalies with 660 abnormal events. The anomalous events are highly challenging, such as violating traffic rules, sleeping, and stealing. It has 29 scenes, each with normal and abnormal videos, which can support supervised training. In our experiments, we only utilized normal videos for training.

\subsection{Implementation Details}
In the pose estimation stage, we first uses AlphaPose \cite{crowdpose} to detect human skeletons and then employs OSNet \cite{reid} to obtain the sequence of human skeletons. During data preprocessing, we apply a moving average window with a size of 2. For the temporal modeling part, we adopt the MAF \cite{maf} model with MADE \cite{made} as the building block, using 3 blocks for ShanghaiTech dataset and 2 blocks for UBnormal dataset. During training, the mini-batch size is set to 256, and the optimizer used is Adamx with a learning rate of 5e-4. The length of the temporal segments L is set to 24 for ShanghaiTech dataset and 16 for UBnormal dataset. Due to space limitations, details of parameter selection cannot be further discussed. \textbf{\textit{The experiments are conducted on a CPU with an Intel(R) Core(TM) i5-10500 CPU @ 3.10GHz.}}


\subsection{Experimental Results}
Tables \ref{ShanghaiTechTable} and \ref{UBnormalTable} provide comparisons between our method and other state-of-the-art approaches on the ShanghaiTech and UBnormal datasets, respectively. The methods compared on ShanghaiTech are all based on skeleton modalities. As we can see, due to the simplicity of our data and model, we cannot compete with state-of-the-art methods in terms of AUC scores. However, it is worth noting that, despite having a model parameter quantity that is \textbf{over four hundred times smaller}, HKVAD-3 achieves comparable AUC performance to MPED-RNN on ShanghaiTech, and trains faster. Moreover, on the most challenging dataset, UBnormal, our method achieves higher AUC scores than even the excellent Jigsaw by 2\% to 3\%, while consuming very limited resources, using only 18, 28, or 38 parameters in the model, which fully demonstrates the effectiveness and potential of our method.

\begin{figure*}[t]
\includegraphics[width=\linewidth]{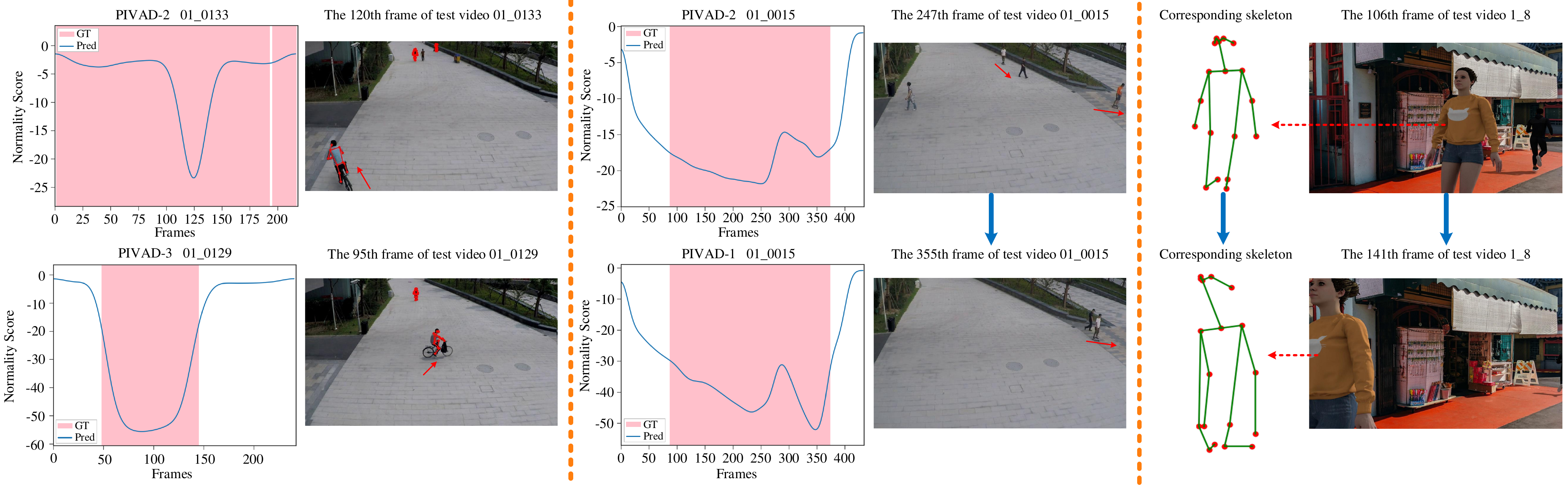}
\caption{The right of each subfigure shows the frames in the dataset, while the left displays the corresponding normality score curve of HKVAD on the corresponding video, or the skeleton data corresponding to the video frame.  Left: visualization of HKVAD for abnormal cycling behavior detection. Middle: comparison of abnormal detection performance between HKVAD-2 and HKVAD-1 on the same video. Right: the detection results of the pose estimator on the video frames.}
\label{discussion}
\end{figure*}

\subsection{Visualization and Analysis}
Despite using basic and unsophisticated information and models, our experimental results demonstrate valuable performance. To better understand the strengths and limitations of our proposed method, we conduct an in-depth analysis in this section.

\noindent
\textbf{How's HKVAD's anomaly detection?}
As shown in the left figure of Figure \ref{discussion}, we can see that when considering the abnormal behavior of pedestrians while cycling, our method can sensitively capture such abnormalities. This is consistent with our intuition that the stride and displacement characteristics exhibited by people while cycling are certainly very different from those during normal walking or standing. Therefore, the successful capture of such abnormalities confirms the rationality of the starting point of the HKVAD methodology.

\noindent
\textbf{What did HKVAD learn?}
In the middle of Figure \ref{discussion}, HKVAD-2 and HKVAD-1 show similar normal score trends on the 01\_0015 video, suggesting that the addition of stride features did not significantly improve the performance of HKVAD-2. This implies that our model can capture simple kinematic features but may not fully understand more complex kinematic features, such as the relationship between displacement and stride in human walking. This limitation may be attributed to the simplicity of the temporal modeling approach we employed.

\noindent
\textbf{Do we need more joints?}
Compared to HKVAD-1 and HKVAD-2, which only use two joints to obtain temporal information, HKVAD-3 additionally utilizes the neck joint to capture the temporal sequence of neck movement. HKVAD-3 outperforms HKVAD-1 and HKVAD-2 on both datasets, indicating that the model's performance improves as more joint data is used (Table). Using more joint data is necessary because the pose estimator's detection results are affected when the person's limbs are not fully visible in the video frame, introducing data noise (right image of Figure \ref{discussion}). Therefore, introducing additional joints can reduce the model's susceptibility to noisy data to some extent. Furthermore, consider a fall action, which cannot be well characterized solely based on data from two foot joints. Thus, to better understand human motion, we need more joint data to extract more kinematic features.

\subsection{Discussion}
From the perspective of data types, the method employed in this paper for extracting kinematic features from skeletal data can be regarded as a means of transforming skeletal data into temporal data. This suggests that it may be worthwhile to analyze the data from both the frequency and time domains, thereby uncovering spatiotemporal patterns. This conceptual approach could potentially be extended to the realm of human skeleton action recognition.
Future work includes: 1) designing more sophisticated models that can learn causal associations between stride-time series and displacement-time series; 2) exploring the use of more joint data and designing more comprehensive human kinematic features to better model spatio-temporal patterns; 3) extending the approach of transforming skeletal data into temporal data for learning spatio-temporal patterns to the field of human skeletal action recognition.

\section{Conclusion}
In this paper, we propose HKVAD, a novel methodology that models human kinematic features to detect anomalies. We have demonstrated the effectiveness and potential of HKVAD by achieving impressive results through experiments conducted on two datasets using three simple variant methods, all within the constraints of limited computational resources. Our work highlights the value of incorporating kinematic features in anomaly detection and provides a promising direction for future research. Additionally, we propose a novel and more challenging anomaly detection scenario where normal events are rare. \textbf{Note:This paper is currently not perfect and there is much work to be done. However, this does not hinder it from providing enlightenment to everyone.} Currently, we are working diligently to develop more suitable modeling approaches.

\bibliography{xiao}
\end{document}